# A Deep Learning-Driven Inhalation Injury Grading Assistant Using Bronchoscopy Images

Yifan Li, Alan W Pang, and Jo Woon Chong

## ABSTRACT

Inhalation injuries present a challenge in clinical diagnosis and grading due to Conventional grading methods such as the Abbreviated Injury Score (AIS) being subjective and lacking robust correlation with clinical parameters like mechanical ventilation duration and patient mortality. This study introduces a novel deep learning-based diagnosis assistant tool for grading inhalation injuries using bronchoscopy images to overcome subjective variability and enhance consistency in severity assessment. Our approach leverages data augmentation techniques, including graphic transformations, Contrastive Unpaired Translation (CUT), and CycleGAN, to address the scarcity of medical imaging data. We evaluate the classification performance of two deep learning models, GoogLeNet and Vision Transformer (ViT), across a dataset significantly expanded through these augmentation methods. The results demonstrate GoogLeNet combined with CUT as the most effective configuration for grading inhalation injuries through bronchoscopy images and achieves a classification accuracy of 97.8%. The histograms and frequency analysis evaluations reveal variations caused by augmentation CUT with distribution changes in the histogram and texture details of the frequency spectrum. PCA visualizations underscore the CUT substantially enhances class separability in the feature space. Moreover, Grad-CAM analyses provide insight into the decision-making process; mean intensity for CUT heatmaps is 119.6, which significantly exceeds 98.8 of the original datasets. Our proposed tool leverages mechanical ventilation periods as a novel grading standard, providing comprehensive diagnostic support.

## Introduction

Thermal injuries are a significant cause of morbidity and mortality in society[1]. Inhalational injuries add to the severity of any thermal injury, and smoke inhalation injuries stand as a critical determinant of patient outcomes [2]. Inhalational injuries add to the severity of any thermal injury. A thermal injury to the oropharynx and a chemical injury to the lungs are often the main components of an inhalational injury [3]. The chemical injury to the lungs is toxic to the pili of the bronchial surface, which are the natural infection-resistance mechanism of evacuating foreign material [4]. This deadly combination leads to a mortality of over 40% in inhalational injuries [5]. In the US, smoke inhalation injuries constituted 7.7% of all burn injury cases in 2019 [6]. The prognosis for burn victims worsens considerably when inhalation injuries are factored in, as evidenced by a marked increase in mortality rates [7]. The decadal span from 2009 to 2018 recorded a mortality rate of 2.9% for burn patients devoid of inhalation injuries, which surged by 20% when such injuries were present. The situation grows grimmer with the addition of pneumonia, where mortality rates skyrocket to 60% [7].

Currently, the most widely used grading system for inhalation injury in clinical practice is the Abbreviated Injury Score (AIS) [8]. This grading scale categorizes the severity of the injury on a scale of "0" (no injury) to "4" (massive injury) based on bronchoscopy imaging. However, Studies have shown an inconsistent cause-and-effect relationship between the inhalation injury grade and the period during which the patient requires mechanical ventilation [9]. The grading of inhalational injury is subjective as a communication tool [10, 11]. The recent study shows a weak correlation between higher-grade inhalational injuries and outcomes such as ventilator days, ICU stays, and other comorbidity, and inhalational injury has a higher incidence as total body surface area increases, making an ill patient even sicker [10]. Additionally, the inhalation injury grade is not highly correlated with mortality [8, 9, 12, 13]. Lack of consensus on diagnosis, grading, and prognosis of inhalation injury stems from the limitations of the AIS grading system, leading to inadequacies in treatment. Bronchoscopy cannot identify narrow distal airway changes, and diagnosis and grading depend on the image quality and interpretation [14]. Recently, a method for grading inhalation injuries based on the duration of mechanical ventilation was introduced[15], in this paper, we proposed a deep learning based grading approach incorporating the ventilation period to build an objective diagnosis system.

Deep learning and artificial intelligence have shown promising medical imaging classification capabilities [16]. Convolutional neural networks (CNNs), a class of artificial neural networks, have been particularly effective for medical image classification and feature recognition [17]. Nevertheless, the scarcity of medical data significantly challenges applying machine learning techniques to bronchoscopy images [18, 19]. Data augmentation has been commonly used in medical imaging for deep learning purposes due to the requirement of larger datasets [18, 20]. With the advances in generative AI, Generative Adversarial Networks (GANs) have been noticed because of their superior image creation skills and are extensively utilized for data augmentation [19, 21]. Auxiliary Classifier GAN (AC-GAN) was also utilized to generate dermoscopic images in order to populate the dataset and alleviate class imbalance[22]. Radiologists evaluated the images of rare malignant tumors generated by GANs with the Likert scale and proved GANs could generate a large number of images from a small sample size [23]. The adversarial generative neural network-based model, Star-GAN, was used for data augmentation on endoscopy images and showed higher performance improvement than the baseline model [24]. Dual encoder variational autoencoder-generative adversarial network (DEVAE-GAN) was proven to be effective in generating high-quality artificial EEG samples.[25] We compare the GAN based models and graphic transformations in this paper, the result suggests that GAN based models are effective with bronchoscopy images.

Obtaining medical data is challenging due to privacy concerns, a lack of annotation experts, under-representation of uncommon conditions, and associated expenses [26, 27]. Transfer learning has been proven effective on multiple tasks, including classification, detection, and segmentation in small clinical datasets [28-32]. Compared with CNNs trained from scratch, finetuned deep CNN models show a more robust sample size and higher performance [32]. For instance, retrained DensNet121 leads to substantial performance improvement (>15% increase on AUC) for small datasets with N<2000 [33]. Hybrid BYOL-ViT obtained a significant boost of performance from 41.66% to 83.25% on the small datasets STL-10 [34]. Tokens-to-Token Vision Transformers (T2T-ViT) combing transfer learning was applied to classify the cervical cancer smear cell image dataset in the liquid-based cytology, achieving Pap smear dataset (4-class), SIPAKMeD (5-class), and Herlev (7-class) are 98.79%, 99.58%, and 99.88 [35]. Transfer learning is also applied in inhalational injury care, GoogLeNet performs 86.11% accuracy in classifying different degrees of injuries [15]. In this paper, we utilize transfer learning and compare performance of different models to figure out optimized approach for determining the severity of inhalation injuries.

With the advancements in AI, Deep Neural Networks (DNN) are widely used for decision-making, but their opacity can hinder trust, especially in crucial domains where understanding decisions is essential. Explainable Artificial Intelligence seeks to interpret these models, with image data posing unique challenges due to class, scale, and background variations [36]. Local Interpretable Model-agnostic Explanations (LIME) approximate complex models locally with simpler models to derive explanations, perturb the input image, observe prediction changes, and fit a simpler model [37]. LIME was applied to explain lightweight and customized convolutional neural network detecting kidney cysts, stones, and tumors, and provides conclusive and understandable results [38]. The Gradient-weighted Class Activation Mapping (Grad-CAM) uses the gradients flowing into the last convolutional layer to produce a heatmap highlighting important regions in the image, providing visual explanations for decisions from CNN-based models [39]. For understanding how augmentation methods working on images, and the features extracted from classification models, we implement both images based, and machine learning based including Grad-CAM analysis in this paper.

We introduce an automated grading system for inhalation injuries using deep learning applied to bronchoscopy images. This system assists in clinical decision-making by providing accurate and objective injury grade estimations, which can guide treatment strategies effectively. By employing advanced data augmentation techniques, our approach overcomes the challenges of imbalanced datasets, typically characterized by a scarcity of diseased versus healthy data. This enhances the model's ability to accurately grade injuries, reducing the risk of overfitting and improving diagnostic reliability. In the context of inhalation injury grading, the combination of well-established methods offers significant advantages tailored to the complexities of our medical problem. The application of data augmentation through GANs addresses the critical issue of limited and imbalanced datasets by generating diverse and high-quality synthetic bronchoscopy images, enhancing the robustness of the training process. Using pre-trained CNNs via transfer learning enables the model to leverage vast amounts of prior knowledge, which is crucial given the specialized nature of bronchoscopy images and the scarcity of annotated medical data. This approach ensures the model can capture the nuanced visual features indicative of varying injury severities. We elucidate how image augmentation influences classification accuracy by Utilizing image-based (histogram and frequency analysis) and machine learning-based (PCA, Grad-CAM) technologies, aligning the model's focus with clinically relevant areas of the bronchial structure. This approach validates the model's enhancements and increases the transparency and interpretability of our deep learning methodology.

## Results

**Data Acquisition and Organization**

In this paper, we delineated a systematic approach to enhancing and analyzing bronchoscopy images from patients with respiratory failure due to burns. The methodology integrates data acquisition, augmentation techniques, and generative models to address the limitations of small datasets. By applying transfer learning to CNN-based and ViT models and comparing their performance, we adapted sophisticated architectures to our specialized dataset, comprising original and augmented images. The flowchart outlines the systematic process employed to develop our inhalation injury diagnosis assistance tool is shown in Figure 1a. Initially, the collected images are allocated into training and testing datasets at a 7:3 ratio. As indicated by the blue pathway in the flowchart, the training set is utilized to develop generative models for data augmentation following graphic augmentation, then the augmented set is fed to finetune both CNN-based and ViT models. The green path represents the testing phase, where the testing set is used to evaluate the performance of the fine-tuned models. To ensure a consistent and fair comparison of model performance, the testing set is not augmented, preserving the original image conditions to reflect real-world application scenarios accurately.

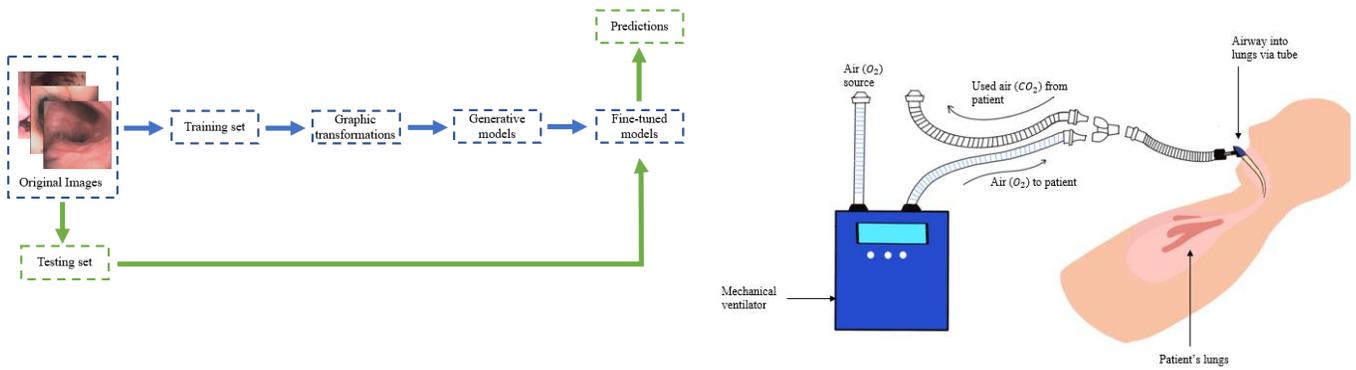

(a) Flowchart depicting the development process of the inhalation injury diagnosis assistance tool. The process begins with bronchoscopy image collection and progresses through data augmentation, model training, and validation.

(b) An example of a mechanical ventilator, the ventilator pulls air and extracts oxygen (O2) from an external source. The patient receives oxygen from the tube passing into the lungs

**Figure 1.** Overview of the proposed approach in this paper

The images utilized in this paper were collected by burn surgeons at the Timothy J. Harnar Regional Burn Center/Department of Surgery at Texas Tech University Health Science Center (TTUHSC), following the institutional review board (IRB) approval (IRB#00000096) for protecting human subjects. Burn critical care physicians assessed the lower airways using a bronchoscope, which included the trachea and bronchi, diagnosing inhalational injury based on visualized carbonaceous deposits, blistering, or fibrin casts. Bronchoscopy with a camera was employed to capture images from patients' bronchi. During image collection, a thin, flexible, tubular camera (bronchoscope) was inserted through the patient's endotracheal tube and into the bronchi, where images of injured bronchi were captured.

To understand the progression and severity of inhalational injuries, patients were categorized into six groups based on the duration of mechanical ventilation. Mechanical ventilators are crucial in treating respiratory failure, especially following inhalational injuries as shown in Figure 1b. These devices assist in delivering oxygen (O2) from an external source to the patient and removing carbon dioxide (CO2) from the lungs. By controlling the rate and volume of ventilation, therapists and doctors optimize the patient's respiratory function. Therefore, the number of days a patient is extubated provides valuable insight into the severity of the inhalational injury and the patient's recovery trajectory.

In this paper, we collected bronchoscopy images from twenty-two patients who got burnt and received ventilation from the mechanical ventilator machine treating respiratory failure. The patient cohort comprised seventeen males and five females with an age range of 18-80 years. Two hundred thirty-six images were collected, with an average of 11 images acquired per patient. Also, we divided the 22 patients into six groups based on the period during which patients received mechanical ventilation: (1) 2 patients were extubating under 24 h as grade 1, (2) 9 patients were extubating between 1–2 days as grade 2, (3) 3 patients were extubating between 3–7 days as grade 3, (4) 3 patients were extubating between 8–14 days as grade 4, (5) 3 patients were extubating between 14–30 days as grade 5, and (6) 2 patients were extubated after 30 days as grade 6.

## Data Augmentation

In the quest to overcome the limitations of a finite and imbalanced dataset, data augmentation plays a pivotal role in amplifying the diversity and volume of available training samples. As illustrated in Table 1, we compare the number of images of different augmentation approaches with the original ones over grades. It shows our augmentation strategies have significantly increased the number of images across all grades of injury, achieving a more balanced distribution across all categories.

| grade   | Original | Transformations | CycleGAN | CUT  |
|---------|----------|-----------------|----------|------|
| grade 1 | 13       | 117             | 702      | 1098 |
| grade 2 | 66       | 385             | 2322     | 837  |
| grade 3 | 40       | 144             | 810      | 1070 |
| grade 4 | 55       | 297             | 1557     | 918  |
| grade 5 | 26       | 117             | 690      | 1098 |
| grade 6 | 36       | 162             | 960      | 1053 |
| total   | 236      | 1222            | 7041     | 6074 |

**Table 1.** Number of Images Before and After Data Augmentation

The augmentation procedure enlarged our dataset and upheld the authenticity of the bronchoscopy images, as seen in the green highlighted section of Figure 1. The synthesized images retain the original's integrity in terms of color, texture, and apparent secretions, confirming that the augmented data preserves the high quality required for practical model training.

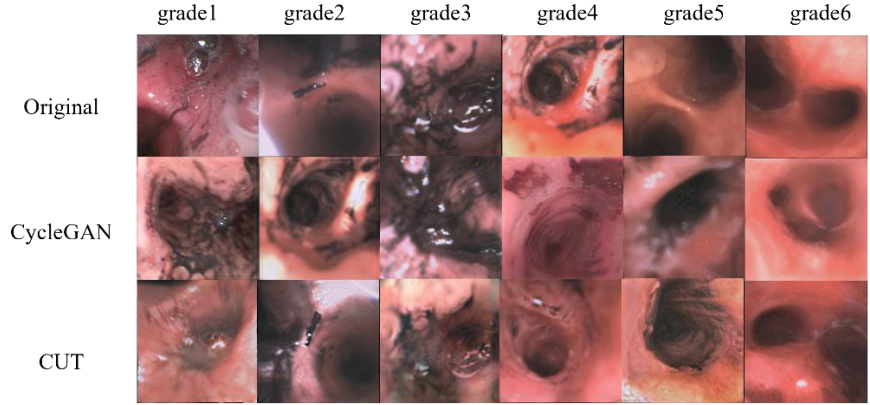

**Figure 1.** Number of Images Before and After Data Augmentation

**Severity Classification**

We utilized GoogLeNet and Vision Transformer (ViT) as classification models to evaluate the severity of inhalation injuries. The performance of these models was assessed based on precision, sensitivity (also known as recall), specificity, accuracy, and the F1 score. These metrics provide a comprehensive overview of the models' ability to classify the severity of inhalation injuries from bronchoscopy images correctly. The results, summarized in TABLE II, indicate the effectiveness of our proposed data augmentation methods in enhancing model performance.

In Table 2, for GoogLeNet, the results demonstrate the CUT method significantly enhances performance, with precision, sensitivity, specificity, accuracy, and F1 scores all exceeding 97%, representing substantial improvements over the original dataset, where the metrics are notably lower (precision at 45.35%, sensitivity at 46.56%, and accuracy at 43.84%). The Transformation method shows a marked improvement over the original dataset, with precision increasing from 45.35% to 87.83% and accuracy from 43.84% to 88.19%, highlighting the efficacy of fundamental transformations in data augmentation. Comparatively, CycleGAN also improves upon the original dataset but does not reach the effectiveness of CUT, with precision and accuracy reaching 88.89%.

| *Augmentation method* | *Precision* | *Sensitivity* | *Specificity* | *Accuracy* | *F1 score* |
|---|---|---|---|---|---|
| **GoogLeNet** | | | | | |
| Original | 0.4535 | 0.4656 | 0.8855 | 0.4384 | 0.4268 |
| Transformations | 0.8783 | 0.8734 | 0.9751 | 0.8819 | 0.8732 |
| CUT | 0.9814 | 0.9792 | 0.9955 | 0.9780 | 0.9794 |
| CycleGAN | 0.8889 | 0.8889 | 0.9778 | 0.8889 | 0.8889 |
| **Vision Transformer (ViT)** | | | | | |
| Original | 0.0525 | 0.1667 | 0.8333 | 0.3152 | 0.0799 |
| Transformations | 0.4277 | 0.4608 | 0.8845 | 0.3973 | 0.3627 |
| CUT | 0.9688 | 0.9690 | 0.9934 | 0.9671 | 0.9685 |
| CycleGAN | 0.7432 | 0.6841 | 0.9547 | 0.7797 | 0.6483 |

**Table 2.** Classification Performance Metrics for GoogLeNet and Vision Transformer (ViT) with Different Augmentation Methods

For ViT, the performance boost with CUT is even more pronounced, with precision jumping from a mere 5.25% in the original dataset to 96.88% and accuracy from 31.52% to 96.71%. The impact of data augmentation on ViT is significant, considering the actual low performance, underscoring the importance of sophisticated augmentation methods like CUT in enhancing model efficacy. CycleGAN's performance, while beneficial, shows a less dramatic improvement in ViT compared to CUT, with accuracy at 77.97%, which is a substantial increase from the original but

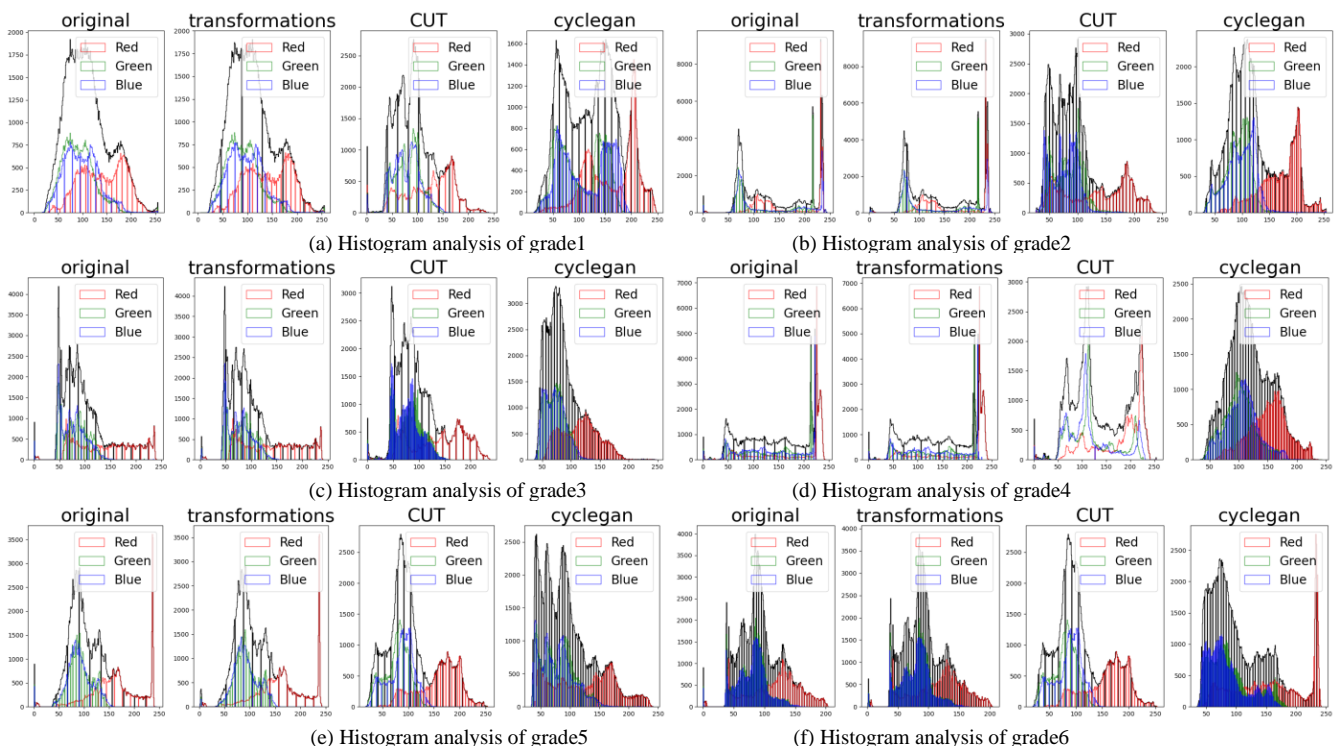

(a) Histogram analysis of grade1  (b) Histogram analysis of grade2
(c) Histogram analysis of grade3  (d) Histogram analysis of grade4
(e) Histogram analysis of grade5  (f) Histogram analysis of grade6

still behind CUT.

The CUT method stands out across both models (ACC: 97.80% in GoogLeNet, 96.71% in ViT), indicating its ability to generate high-quality images that improve model training and performance. The difference in performance metrics between the original and augmented datasets, especially with CUT (i.e, CUT improves 53.96% accuracy with GoogLeNet and 65.19% with ViT), underscores the critical role of augmentation technologies in overcoming the limitations of small and imbalanced datasets.

**Result Interpretation**

*Histogram analysis*

Figure 2 informs us about pixel intensity variations which are crucial for image classification. Each injury grade's histogram reveals distinct changes post-augmentation, influencing classifier performance. In Figure 2, black line in the histogram represents overall intensity of images, which means the distribution of all pixel intensities across all color channels combined, red line represents the distribution of pixel intensities in the red channel of images. It highlights the presence and variation of the red channel, similarly, the blue and green lines map out the intensity distributions for the blue and green channels.

**Figure 2.** Histogram analysis showcasing the distribution characteristics of the original and generated images

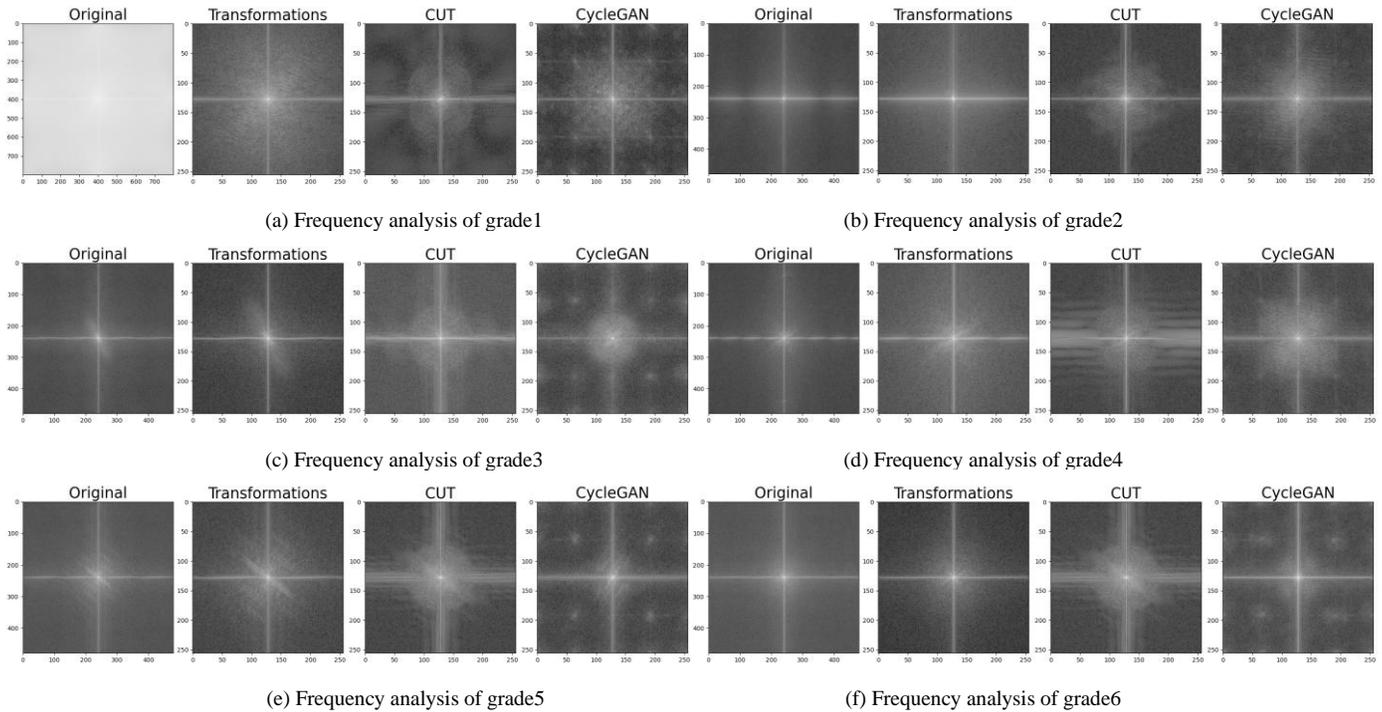

(a) Frequency analysis of grade1

(b) Frequency analysis of grade2

(c) Frequency analysis of grade3

(d) Frequency analysis of grade4

(e) Frequency analysis of grade5

(f) Frequency analysis of grade6

Original images across grades typically show an even intensity spread. CUT and CycleGAN methods further accentuate this trend by sharpening contrast and enhancing feature visibility. For instance, CUT often generates a distinct peak in the lower intensity range, implying greater contrast and detail recognition potential.

Across all grades, augmentation consistently heightens contrast, particularly in darker features, aiding classifiers in feature discrimination. The CUT method presents more defined intensity peaks that promise improved model learning and recognition capabilities.

*Frequency analysis*

Frequency analysis, shown in Figure 3, delves into how image augmentations impact the presence of low and high-frequency details, which are integral to classification accuracy. In the frequency spectrum of an image, the central region represents the low-frequency components (in the range from 180-300 in the original dataset and 100-150 in the transformed, CUT, and CycleGAN generated dataset), which correspond to the structures and gradual intensity variations in the image. As moving outward from the center, there are higher frequency components (in the range from 0-180 and 300-450 in the original dataset and 0-100 and 150-256 in the transformed, CUT, and CycleGAN generated dataset), corresponding to details such as edges, textures, and granular patterns.

**Figure 3.** Frequency analysis showcasing the frequency characteristics of the original and generated images. These analyses elucidate the reasons behind the improved classification

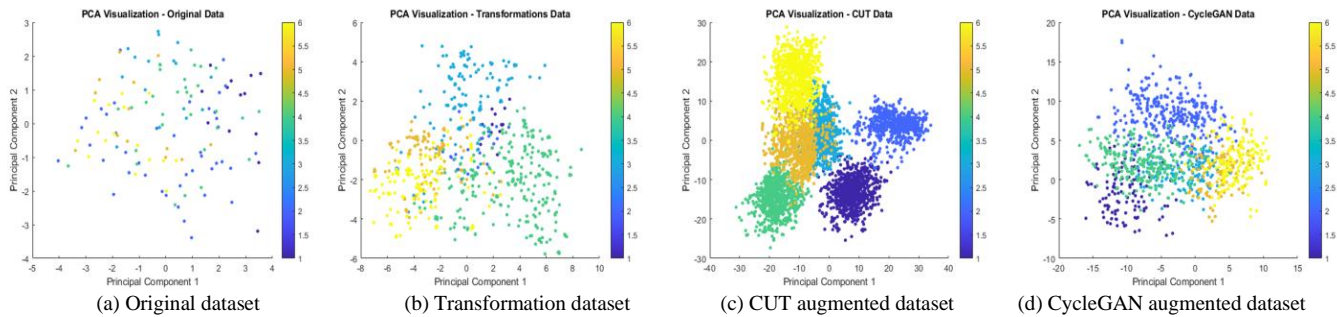

(a) Original dataset    (b) Transformation dataset    (c) CUT augmented dataset    (d) CycleGAN augmented dataset

performance of the generated images

Throughout the grades, the augmentation methods introduce improvements in frequency detail. CUT's method increases brightness in high-frequency areas, translating to better-defined details. CycleGAN ensures an even spectrum representation, reinforcing basic structures and fine details.

The observed improvements in frequency distribution through the CUT and CycleGAN augmentations hint at their potential to augment classifier performance by providing richer model training and interpretation features.

### *PCA (Principal component analysis) feature visualization*

Principal Component Analysis (PCA) is utilized to reduce the dimensions of the feature space extracted from the images, enabling a visual inspection of the data structure in a two-dimensional plot. To analyze the separability of the image classes, feature vectors were extracted from the penultimate fully connected layer of the GoogLeNet architecture. This layer is known for its rich, abstract representations of the input data, which are particularly suitable for visualizing and understanding the clustering of image classes. The extracted features were then subjected to Principal Component Analysis (PCA) to reduce dimensionality and visualize the data in a two-dimensional plot. The PCA plots for different data processing techniques are illustrated in Figure 4, providing insight into the class separation achieved by each method.

**Figure 4.** PCA-visualization for different data augmentation method. The improved separability suggests the data augmentation phase effectively improve the clustering of image classes.

The PCA visualization of the original data, as shown in Figure 4a, reveals a scattered distribution of feature vectors without clear boundaries. This dispersion indicates substantial overlap in the feature space, making it challenging for classifiers to differentiate between classes accurately.

In contrast, Figure 4b displays PCA plot for the transformations data showing the initial formation of clusters, although considerable intermixing persists. This suggests that while transformations begin to enhance class separability, the distinction between classes is not yet well-defined.

Visualizing the CUT data in Figure 4c presents a marked improvement, with feature vectors forming distinct, well-separated clusters. Such clear demarcation between classes implies that the feature vectors derived from the CUT-processed images possess unique characteristics that are more readily identifiable, potentially improving classifier accuracy.

Lastly, the PCA visualization for CycleGAN data shown in Figure 4d depicts clusters that, while not as sharply defined as those from the CUT method, exhibit improved separability over the

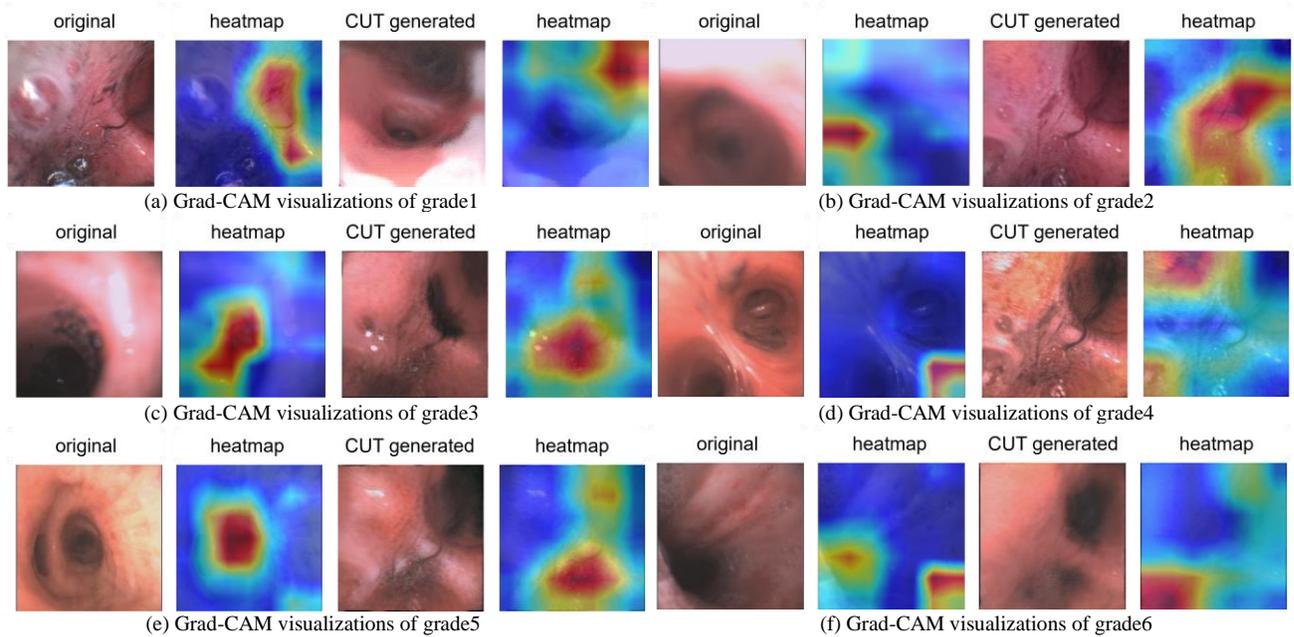

(a) Grad-CAM visualizations of grade1
(b) Grad-CAM visualizations of grade2
(c) Grad-CAM visualizations of grade3
(d) Grad-CAM visualizations of grade4
(e) Grad-CAM visualizations of grade5
(f) Grad-CAM visualizations of grade6

original dataset. This enhanced separability indicates that CycleGAN effectively balances preserving essential features with introducing class-distinctive attributes.

Generated images exhibiting greater divisibility in feature space are essential for classifier efficacy. Convolutional neural networks and other image classification models depend on distinguishable patterns within the feature space to well-defined clusters, which is advantageous for classification tasks. The observations from these visualizations support the premise that data augmentation via CUT and CycleGAN can significantly enhance the learning process by producing more distinctive and classifiable feature representations.

**Gradient-weighted Class Activation Mapping (Grad-CAM) visualization**

Grad-CAM offers visual explanations for decisions made by convolutional neural networks, highlighting the influential regions in images that lead to classification decisions. By overlaying heatmaps on the original and generated images, Grad-CAM provides insight into the network's focus areas during the classification process.

Our observations from the Grad-CAM heatmaps, as shown in Figure 5, reveal that the network predominantly focuses on the bronchial holes for the original images, particularly for grades 3, 5, and 6 (see in Figure 5c, 8e, and 8f). This fixation poses a significant challenge in accurately classifying the severity of inhalation injuries, as the condition's true indicators might reside elsewhere in the bronchial structure.

**Figure 5.** CAM visualizations indicating the focus areas of GoogLeNet on both original and

CUT-generated images across different grades of injury. The heatmaps elucidate the shift in concentration from bronchial holes to the bronchus's inner wall, signifying an improved alignment with clinically relevant features for classification.

In contrast, the heatmaps for the CUT-generated images indicate a shift in the network's attention towards the inner walls of the bronchus. This shift suggests that the generated images can guide the network to concentrate on more relevant features for assessing the injury severity. Moreover, it is noteworthy that the network does not seem distracted by any noise introduced during the CUT augmentation process. For instance, the heatmap for grade 1 of the CUT group does not highlight the peripheral noise shown as white areas seen the CUT generated image in Figure 5a, confirming that the network's focus remains on the salient features within the image.

The mean intensity of the pixels in the Grad-CAM heatmaps serves as a quantitative measure of the model's focus across the image. Higher mean intensity values suggest that the model perceives the features within the image as more distinguishable, which is advantageous for classification tasks. In this paper, we calculated the mean intensity for each grade of injury across different image generation methods to ascertain how these methods affect the model's attention.

As summarized in Table 3, the CUT method generally exhibits higher mean intensities than the original images, implying that the augmented images from this method might enhance features critical for classification. This enhancement is consistent across different grades, signifying a systematic improvement in feature presentation. It is particularly noteworthy in Grade 4, where the mean intensity for CUT stands out, indicating a substantial increase in model attentiveness to features relevant to classifying this grade of injury.

| grade | Original | Transformations | CycleGAN | CUT |
|---|---|---|---|---|
| grade 1 | 96.47 | 112.8 | 103.72 | 112.4 |
| grade 2 | 120.35 | 90.36 | 118.5 | 124.26 |
| grade 3 | 110.58 | 117.54 | 111.15 | 119.13 |
| grade 4 | 71.22 | 120.07 | 113.79 | 131.43 |
| grade 5 | 110.21 | 117.88 | 87.79 | 117.09 |
| grade 6 | 92.53 | 101.67 | 87.79 | 114.41 |

**Table 3.** CAM visualizations indicating the focus areas of GoogLeNet on both original and CUT-generated images across different grades of injury. The heatmaps elucidate the shift in concentration from bronchial holes to the bronchus's inner wall, signifying an improved alignment with clinically relevant features for classification.

## Discussion

Our comprehensive study on inhalation injury grading using deep learning presents several significant findings. Data augmentation techniques, including graphic transformations, CycleGAN, and CUT, have substantially expanded our dataset, ensuring ample variability for model training. The effectiveness of these methods is evident in the improved performance metrics of GoogLeNet and Vision Transformer models, with the CUT method standing out, as seen in Table 2. The Contrastive Unpaired Translation (CUT) method has notably improved classification metrics across both models. For GoogLeNet, metrics such as precision, sensitivity, specificity, accuracy, and F1 score have increased with applying CUT, with all metrics over 97%, compared with the original dataset (precision at 45.35%, sensitivity at 46.56%, and accuracy at 43.84%). The application of graphic transformations also resulted in improvements in performance metrics, the precision and accuracy metrics for GoogLeNet improved from 45.35% to 87.83% and from 43.84% to 88.19%, respectively. CycleGAN also achieved improvements, with precision and accuracy reaching 88.89%. For the ViT model, applying the CUT method increased precision from 5.25% to 96.88% and accuracy from 31.52% to 96.71%. CycleGAN also contributed to improved performance with an accuracy of 77.97%. The differential in performance metrics between the original and CUT-augmented datasets emphasizes the importance of proposed data augmentation for addressing the problem of small and imbalanced datasets in inhalation injury grading. Both GoogLeNet and ViT models substantially benefit from the diversity introduced by the CUT method. More accurate and reliable grading of inhalation injuries is crucial for assisting clinical decision-making and optimizing patient care.

Histogram analysis in Figure 2 shows how pixel intensity distributions in bronchoscopy images across grades 1 to 6 change post-augmentation, affecting the ability to differentiate injury severity. The CUT method notably enhances contrast across all grades, especially at the intensity range extremes, while CycleGAN introduces broader intensity variations. Grade-specific findings show that fundamental transformations preserve the original intensity profile for grade 1, whereas CUT and CycleGAN significantly change it. For grades 2 and 3, these augmentations increase the intensity range, improving feature visibility. Grades 4 and 5 shift towards higher mid-intensity values with CUT, highlighting features crucial for higher injury severity classification. Grade 6 continues this pattern, with CUT amplifying mid-to-high intensity values in the red channel, aiding in severe injury recognition. Overall, histograms of original images have a balanced intensity distribution, but CUT often introduces distinct low-intensity peaks, enhancing contrast and detail detection. CUT consistently enhances contrast, especially in darker features, aiding classifiers in feature discrimination. These findings emphasize the importance of tailored data augmentation strategies, with CUT and CycleGAN improving image quality for classification and enriching feature understanding, contributing to more precise inhalation injury grading.

Frequency analysis (Figure 3) shows augmentation techniques impact textural features for inhalation injury grading. In grades 1 through 6, CUT and CycleGAN augmentations significantly increase low- and high-frequency details, suggesting enriched textural features suitable for classification tasks. For example, grade 1 images lack high-frequency detail, but CUT augmentation increases detail across the spectrum. Grades 2 through 6 demonstrate similar trends, with augmented images showing rising textural detail and complexity. The spectral enhancements from lower to higher grades indicate an overall improvement in textural information for the classifier. Moreover, the augmentation methods retain and emphasize necessary original image features. Overall, frequency analysis highlights the effectiveness of data augmentation techniques in improving classifier performance by introducing crucial texture features and retaining the structure of the original images.

PCA analysis shows that augmentation techniques, particularly CUT, significantly improve class separability in the feature space. PCA plots from GoogLeNet's fully connected layer reveal that the original dataset lacks distinct feature boundaries, causing overlap and inaccurate classification (Figure 4a). Transformation methods offer some clustering but fail to provide clear separation (Figure 4b). In contrast, CUT and CycleGAN create distinct, non-overlapping clusters (Figs. 7c and 7d), crucial for classifier performance as it ensures unique class-specific features for accurate predictions. Data augmentation, especially with CUT, creates more defined feature boundaries, enhancing the inhalation injuries grading system's performance.

Grad-CAM analyses reveal how the convolutional neural network focuses on image regions for classification. Original images highlight bronchial holes that lack the necessary information for accurate classification of injury severity (Figure 5b, c, e, and f). The CUT method redirects focus towards diagnostically relevant inner bronchus walls, reducing noise (Figure 5a) and indicating improved image quality and clinical relevance. Quantitative analysis with mean intensity values from Grad-CAM heatmaps across grades confirms these observations, particularly in Grade 4, where CUT images show significantly higher mean intensity than the original dataset (Table 3).

The novel approach of our inhalation injury grading method, which is anchored on the duration of mechanical ventilation, presents a shift from traditional AIS-based assessments. Our analysis suggests a more objective classification, correlating the injury severity with the patient's reliance on mechanical ventilation. By incorporating the duration of mechanical ventilation into our data augmentation and machine learning framework as the baseline, we will be able to ensure that the models fit significant features, and potentially enhance their predictive performance and clinical utility.

In conclusion, our findings advocate for incorporating sophisticated data augmentation techniques in medical image analysis, potentially improving the accuracy and reliability of automated grading systems significantly. The innovative diagnostic tool we have developed exemplifies the profound capacity of AI to assist burn surgeons by utilizing the duration of mechanical ventilation as a novel standard for assessing inhalation injury severity. Beyond mere image analysis, our multifaceted deep learning approach provides a more comprehensive diagnostic support system. The comparative performance analysis demonstrates GoogLeNet combined with Contrastive Unpaired Translation (CUT) as the most effective configuration for grading inhalation injuries through bronchoscopy images. For the next step, we aim to enhance our models with additional imaging modalities and to operationalize them in clinical settings. Such initiatives promise to narrow the divide between technological advances and their medical applications, facilitating more accurate, informed, and superior patient care.

## Methods

### Data Augmentation

We employed graphic transformations and Generative Adversarial Network (GAN)-based models to augment our dataset to overcome the constraints imposed by limited dataset size. Further details are provided below:

### *Graphic Transformations*

We utilized graphic transformations on the bronchoscopy images to enrich our dataset before employing generative models and ensuring diverse input. This step aimed to simulate various possible real-world scenarios and variations that might not be present in the initial dataset.

Each bronchoscopy image was first resized to a standard resolution of 256×256 pixels to maintain uniformity across the dataset. Following this, a series of transformations were applied:

- Scaling: Scaling was implemented to simulate potential variations in the images' capturing distances or zoom levels. Images were scaled independently along the x-axis and y-axis by a random factor from the range of [1.1,1.5]. Furthermore, a combined scaling was performed where images were scaled along both axes concurrently.

- Rotation: Considering the various orientations in which bronchoscopy images might be captured, each image underwent rotations at three distinct angles: 90˚, 180˚, and 270˚.

- Reflection: To create a more comprehensive dataset by including possible variations in image orientation, each image was reflected along the x-axis and y-axis.

- Cropping: Cropping is particularly effective in our context, as the bronchoscopy images lack distinct shapes and most cropped sections can adequately represent the condition of the bronchial mucosa

Post-transformation, each resultant image was resized to 256×256 pixels, ensuring a consistent resolution. These augmented images were stored alongside the originals, expanding the breadth and diversity of our dataset before it was fed into the generative models.

### *Generative models*

Generative models are a powerful tool to augment and transform datasets, especially when dealing with limited or imbalanced data. In this paper, we utilized the capabilities of generative models to enhance the diversity and richness of our bronchoscopy image dataset. We employed two generative models, CycleGAN [40] and Contrastive Unpaired Translation (CUT) [41]. These models were chosen for their ability to perform image-to-image translations without needing paired training examples. Our aim was not only to generate diverse images but also to compare the performance of these models to identify the most optimized approach for our dataset.

- Data partitioning: One-vs-All approach: To structure the dataset optimally for generative modeling on our multi-class problem, we implemented a one-vs-all partitioning approach as shown in Fig. 2. For each injury grade, images corresponding to that grade were treated as the target domain (trainB). Conversely, images from all other grades were collectively treated as the source domain (trainA). For instance, for the grade 1 partition, images of grade 1 constitute the target domain, while images from grade 2 to grade 6 collectively form the source domain. This approach ensures the model learns the transformation between a specific injury grade and all others.

- Cycle-Consistent Adversarial Network (CycleGAN): CycleGAN is a popular

unsupervised image-to-image translation method. It enables the transformation of images from one domain to another without paired training examples. The primary components of CycleGAN include two generator networks $G$ and $F$, and two discriminator networks $D_X$ and $D_Y$. The generators are responsible for translating images from one domain to another and vice versa, while the discriminators aim to distinguish between real and generated images. The objective function for CycleGAN includes an adversarial loss and a cycle-consistency loss. The adversarial loss is defined in (1):

$$\mathcal{L}_{GAN}(G, D_Y, X, Y) = \mathbb{E}_{y \sim P_{data(y)}}[log D_Y(y)] + \mathbb{E}_{x \sim P_{data(x)}}[log(1 - D_Y(G(x)))] \quad (1)$$

The cycle-consistency loss in (2) ensures that when an image is translated from one domain to the other and then back, it remains unchanged:

$$\mathcal{L}_{cyc}(G, F) = \mathbb{E}_{x \sim P_{data(x)}}[\|F(G(x)) - x\|_1] - \mathbb{E}_{y \sim P_{data(y)}}[\|G(F(y)) - y\|_1] \quad (2)$$

- Contrastive Unpaired Translation (CUT): CUT integrates contrastive learning into the unpaired image-to-image translation paradigm. The core idea behind CUT is to enforce patch-wise contrastive losses to ensure that local regions in the generated image match the local regions in the actual image from the target domain. The contrastive loss in CUT is given by (3):

$$\mathcal{L}_{CUT}(G, D_Y, X, Y) = -log(\frac{exp(sim(G(x), y_+)/\tau}{exp(sim(G(x), y_+)/\tau) + \sum_{y_-} exp(sim(G(x), y_-)/\tau))}) \quad (3)$$

where $sim(a, b)$ denotes the similarity between patches a and b, $y_+$ is the positive sample from the target domain, and $y_-$ are the negative samples. $\tau$ is a temperature parameter that scales the similarities.

**Injuries Grades Classification**

In this paper, we employed two distinct architectures for image classification: Convolutional Neural Networks (CNNs) and Vision Transformers (ViT) [42]. CNNs, with their hierarchical structure, are adept at capturing local and incremental image features and have established themselves as a cornerstone in image recognition. On the other hand, Vision Transformers utilize self-attention mechanisms to process images like natural language processing, considering the entire context of the image patches to capture global dependencies.

To harness the extensive knowledge these models have acquired from large-scale datasets, we utilized the technique of transfer learning. Transfer learning allows us to fine-tune pre-trained models to our specific dataset, enhancing model performance in tasks with limited data. The underlying principle of transfer learning can be represented in (4)

$$f_{new}(x) = \phi(f_{pre-trained}, \theta_{new}) \quad (4)$$

where $f_{pre-trained}$ denotes the feature extraction function learned from the original large dataset, $\phi$ is the adaptation function fitting the pretrained model with our dataset, $\theta_{new}$ represents the parameters fine-tuned on our dataset, and $f_{new}$ is the resulting model adapted to our classification task.

For our CNN-based model, we initiated with architectures that have been pre-trained on extensive image datasets. We then fine-tuned the final layers where the high-level reasoning takes place, allowing the network to tailor its understanding to the features specific to bronchoscopy images. Similarly, for the ViT model, we adapted the pre-trained transformer by fine-tuning its layers to understand the spatial hierarchies and patterns pertinent to our medical images.

## References

[1]	C. E. Güldoğan, M. Kendirci, E. Gündoğdu, and A. Ç. Yastı, "Analysis of factors associated with mortality in major burn patients," *Turkish journal of surgery,* vol. 35, no. 3, p. 155, 2019.
[2]	A. B. Association, "National Burn Repository: 2002 Report Dataset version 8."
[3]	K. Gupta, M. Mehrotra, P. Kumar, A. R. Gogia, A. Prasad, and J. A. Fisher, "Smoke inhalation injury: etiopathogenesis, diagnosis, and management," *Indian journal of critical care medicine: peer-reviewed, official publication of Indian Society of Critical Care Medicine,* vol. 22, no. 3, p. 180, 2018.
[4]	G. Foncerrada *et al.*, "Inhalation injury in the burned patient," *Annals of plastic surgery,* vol. 80, no. 3 Suppl 2, p. S98, 2018.
[5]	R. El-Helbawy and F. Ghareeb, "Inhalation injury as a prognostic factor for mortality in burn patients," *Annals of burns and fire disasters,* vol. 24, no. 2, p. 82, 2011.
[6]	A. B. Association, "National Burn Repository: 2019 Update Dataset version 14.0 " 2019. [Online]. Available: https://sk75w2kudjd3fv2xs2cvymrg-wpengine.netdna-ssl.com/wp-content/uploads/2020/05/2019-ABA-Annual-Report_FINAL.pdf.
[7]	A. Veeravagu *et al.*, "National trends in burn and inhalation injury in burn patients: results of analysis of the nationwide inpatient sample database," *Journal of Burn Care & Research,* vol. 36, no. 2, pp. 258-265, 2015.
[8]	F. W. Endorf and R. L. Gamelli, "Inhalation injury, pulmonary perturbations, and fluid resuscitation," *Journal of burn care & research,* vol. 28, no. 1, pp. 80-83, 2007.
[9]	S. W. Jones, F. N. Williams, B. A. Cairns, and R. Cartotto, "Inhalation injury: pathophysiology, diagnosis, and treatment," *Clinics in plastic surgery,* vol. 44, no. 3, pp. 505-511, 2017.
[10]	P. F. Walker *et al.*, "Diagnosis and management of inhalation injury: an updated review," *Critical Care,* vol. 19, no. 1, pp. 1-12, 2015.
[11]	S. Tanizaki, "Assessing inhalation injury in the emergency room," *Open Access Emergency Medicine,* pp. 31-37, 2015.
[12]	M. J. Mosier, T. N. Pham, D. R. Park, J. Simmons, M. B. Klein, and N. S. Gibran, "Predictive value of bronchoscopy in assessing the severity of inhalation injury," *Journal of burn care & research,* vol. 33, no. 1, pp. 65-73, 2012.
[13]	J. M. Albright *et al.*, "The acute pulmonary inflammatory response to the graded severity of smoke inhalation injury," *Critical care medicine,* vol. 40, no. 4, p. 1113, 2012.
[14]	P. Enkhbaatar *et al.*, "Pathophysiology, research challenges, and clinical management of smoke inhalation injury," *The Lancet,* vol. 388, no. 10052, pp. 1437-1446, 2016.
[15]	Y. Li *et al.*, "Inhalation Injury Grading Using Transfer Learning Based on Bronchoscopy Images and Mechanical Ventilation Period," *Sensors,* vol. 22, no. 23, p. 9430, 2022.
[16]	M. A. Mazurowski, M. Buda, A. Saha, and M. R. Bashir, "Deep learning in radiology: An overview of the concepts and a survey of the state of the art with focus on MRI," *Journal of magnetic resonance imaging,* vol. 49, no. 4, pp. 939-954, 2019.
[17]	R. Yamashita, M. Nishio, R. K. G. Do, and K. Togashi, "Convolutional neural networks: an overview and application in radiology," *Insights into imaging,* vol. 9, no. 4, pp. 611-629, 2018.

[18] P. Chlap, H. Min, N. Vandenberg, J. Dowling, L. Holloway, and A. Haworth, "A review of medical image data augmentation techniques for deep learning applications," *Journal of Medical Imaging and Radiation Oncology,* vol. 65, no. 5, pp. 545-563, 2021.
[19] Y. Chen *et al.*, "Generative adversarial networks in medical image augmentation: A review," *Computers in Biology and Medicine,* vol. 144, p. 105382, 2022.
[20] Z. Hussain, F. Gimenez, D. Yi, and D. Rubin, "Differential data augmentation techniques for medical imaging classification tasks," in *AMIA annual symposium proceedings*, 2017, vol. 2017: American Medical Informatics Association, p. 979.
[21] A. Biswas *et al.*, "Generative adversarial networks for data augmentation," in *Data Driven Approaches on Medical Imaging*: Springer, 2023, pp. 159-177.
[22] M. Alsaidi, M. T. Jan, A. Altaher, H. Zhuang, and X. Zhu, "Tackling the class imbalanced dermoscopic image classification using data augmentation and GAN," *Multimedia Tools and Applications,* pp. 1-27, 2023.
[23] S. Mulé *et al.*, "Generative adversarial networks (GAN)-based data augmentation of rare liver cancers: The SFR 2021 Artificial Intelligence Data Challenge," *Diagnostic and Interventional Imaging,* vol. 104, no. 1, pp. 43-48, 2023.
[24] H.-C. Park, I.-P. Hong, S. Poudel, and C. Choi, "Data Augmentation based on Generative Adversarial Networks for Endoscopic Image Classification," *IEEE Access,* 2023.
[25] C. Tian, Y. Ma, J. Cammon, F. Fang, Y. Zhang, and M. Meng, "Dual-encoder VAE-GAN with spatiotemporal features for emotional EEG data augmentation," *IEEE Transactions on Neural Systems and Rehabilitation Engineering,* 2023.
[26] F. Mahmood, R. Chen, S. Sudarsky, D. Yu, and N. J. Durr, "Deep learning with cinematic rendering: fine-tuning deep neural networks using photorealistic medical images," *Physics in Medicine & Biology,* vol. 63, no. 18, p. 185012, 2018.
[27] Y. Yu, H. Lin, J. Meng, X. Wei, H. Guo, and Z. Zhao, "Deep transfer learning for modality classification of medical images," *Information,* vol. 8, no. 3, p. 91, 2017.
[28] N. Tajbakhsh *et al.*, "Convolutional neural networks for medical image analysis: Full training or fine tuning?," *IEEE transactions on medical imaging,* vol. 35, no. 5, pp. 1299-1312, 2016.
[29] A. Kumar, J. Kim, D. Lyndon, M. Fulham, and D. Feng, "An ensemble of fine-tuned convolutional neural networks for medical image classification," *IEEE journal of biomedical and health informatics,* vol. 21, no. 1, pp. 31-40, 2016.
[30] G. Wang *et al.*, "Interactive medical image segmentation using deep learning with image-specific fine tuning," *IEEE transactions on medical imaging,* vol. 37, no. 7, pp. 1562-1573, 2018.
[31] Z. N. K. Swati *et al.*, "Brain tumor classification for MR images using transfer learning and fine-tuning," *Computerized Medical Imaging and Graphics,* vol. 75, pp. 34-46, 2019.
[32] N. Tajbakhsh *et al.*, "On the necessity of fine-tuned convolutional neural networks for medical imaging," *Deep Learning and Convolutional Neural Networks for Medical Image Computing: Precision Medicine, High Performance and Large-Scale Datasets,* pp. 181-193, 2017.
[33] M. Romero, Y. Interian, T. Solberg, and G. Valdes, "Targeted transfer learning to improve performance in small medical physics datasets," *Medical physics,* vol. 47, no. 12, pp. 6246-6256, 2020.
[34] S. Naimi, R. van Leeuwen, W. Souidene, and S. B. Saoud, "Hybrid BYOL-ViT: Efficient approach to deal with small datasets," *arXiv preprint arXiv:2111.04845,* 2021.
[35] C. Zhao, R. Shuai, L. Ma, W. Liu, and M. Wu, "Improving cervical cancer classification with imbalanced datasets combining taming transformers with T2T-ViT," *Multimedia tools and applications,* vol. 81, no. 17, pp. 24265-24300, 2022.
[36] L. K. Gupta, D. Koundal, and S. Mongia, "Explainable methods for image-based deep learning: a review," *Archives of Computational Methods in Engineering,* vol. 30, no. 4, pp. 2651-2666, 2023.
[37] M. T. Ribeiro, S. Singh, and C. Guestrin, "" Why should i trust you?" Explaining the predictions of any classifier," in *Proceedings of the 22nd ACM SIGKDD international conference on knowledge discovery and data mining*, 2016, pp. 1135-1144.
[38] M. Bhandari, P. Yogarajah, M. S. Kavitha, and J. Condell, "Exploring the Capabilities of a Lightweight CNN Model in Accurately Identifying Renal Abnormalities: Cysts, Stones, and Tumors, Using LIME and SHAP," *Applied Sciences,* vol. 13, no. 5, p. 3125, 2023.
[39] R. R. Selvaraju, M. Cogswell, A. Das, R. Vedantam, D. Parikh, and D. Batra, "Grad-cam: Visual explanations from deep networks via gradient-based localization," in *Proceedings of the IEEE international conference on computer vision*, 2017, pp. 618-626.

[40] J.-Y. Zhu, T. Park, P. Isola, and A. A. Efros, "Unpaired image-to-image translation using cycle-consistent adversarial networks," in *Proceedings of the IEEE international conference on computer vision*, 2017, pp. 2223-2232.

[41] T. Park, A. A. Efros, R. Zhang, and J.-Y. Zhu, "Contrastive learning for unpaired image-to-image translation," in *Computer Vision–ECCV 2020: 16th European Conference, Glasgow, UK, August 23–28, 2020, Proceedings, Part IX 16*, 2020: Springer, pp. 319-345.

[42] A. Dosovitskiy *et al.*, "An image is worth 16x16 words: Transformers for image recognition at scale," *arXiv preprint arXiv:2010.11929,* 2020.